\documentclass[letterpaper]{article} % DO NOT CHANGE THIS
\usepackage{aaai24}  % DO NOT CHANGE THIS
\usepackage{times}  % DO NOT CHANGE THIS
\usepackage{helvet}  % DO NOT CHANGE THIS
\usepackage{courier}  % DO NOT CHANGE THIS
\usepackage[hyphens]{url}  % DO NOT CHANGE THIS
\usepackage{graphicx} % DO NOT CHANGE THIS
\usepackage{natbib}  % DO NOT CHANGE THIS AND DO NOT ADD ANY OPTIONS TO IT
\usepackage{caption} % DO NOT CHANGE THIS AND DO NOT ADD ANY OPTIONS TO IT
\usepackage{amssymb}
\usepackage{amsmath}
\usepackage{amsthm}
\usepackage{color}
\usepackage{comment}
\usepackage{graphicx}
\usepackage{float}
\usepackage{subfigure}
\usepackage{booktabs}
\usepackage{multirow}
\usepackage{algorithm}
\usepackage{algorithmic}
\usepackage{newfloat}
\usepackage{listings}
\DeclareCaptionStyle{ruled}{labelfont=normalfont,labelsep=colon,strut=off} % DO NOT CHANGE THIS
\floatstyle{ruled}
\newfloat{listing}{tb}{lst}{}
\floatname{listing}{Listing}
\title{Federated Causality Learning with Explainable Adaptive Optimization}
\author{
  Dezhi Yang\textsuperscript{\rm 1,2},
   Xintong He\textsuperscript{\rm 3},
  Jun Wang\textsuperscript{\rm 2,}\thanks{Corresponding author.},
    Guoxian Yu\textsuperscript{\rm 1,2},
    Carlotta Domeniconi\textsuperscript{\rm 4},
    Jinglin Zhang\textsuperscript{\rm 5}
}
\affiliations{
    \textsuperscript{\rm 1}School of Software, Shandong University, Jinan, China\\
    \textsuperscript{\rm 2}SDU-NTU Joint Centre for AI Research, Shandong University, Jinan, China\\
     \textsuperscript{\rm 3}Department of Mathematics, National University of Singapore, Singapore\\
    \textsuperscript{\rm 4}Departiment of Computer Science, George Mason University, VA, USA\\
    \textsuperscript{\rm 5}School of Control Science and Engineering, Shandong University, Jinan, China\\
    dzyang@mail.sdu.edu.cn, E0966399@u.nus.edu.sg, \{kingjun, gxyu,jinglin.zhang\}@sdu.edu.cn, carlotta@cs.gmu.edu
}

\begin{document}

\maketitle

\begin{abstract}
 Discovering the causality from observational data is a crucial task in various scientific domains. With increasing awareness of privacy, data are not allowed to be exposed, and it is very hard to learn causal graphs from dispersed data, since these data may have different distributions. In this paper, we propose a federated causal discovery strategy (FedCausal) to learn the unified global causal graph from decentralized heterogeneous data. We design a global optimization formula to naturally aggregate the causal graphs from client data and constrain the acyclicity of the global graph without exposing local data. Unlike other federated causal learning algorithms, FedCausal unifies the local and global optimizations into a complete directed acyclic graph (DAG) learning process with a flexible optimization objective. We prove that this optimization objective has a high interpretability and can adaptively handle homogeneous and heterogeneous data. Experimental results on synthetic and real datasets show that FedCausal can effectively deal with non-independently and identically distributed (non-iid) data and has a superior performance.
\end{abstract}

\section{Introduction}

Causal structure discovery is a fundamental and critical problem in many fields, such as economics \cite{koller2009probabilistic} and biology \cite{sachs2005causal}. Randomized controlled experiments are the golden standard for discovering causal structure, but they are often limited by cost and may even be prohibited by ethics. Causal discovery typically infers a directed acyclic graph (DAG) from observational data at a central site \cite{pearl2009causality,peters2017elements};  this DAG encodes causal relationships between variables. However, with the increasing awareness of privacy and security, data are scattered among different clients and cannot be shared, which makes it difficult for canonical causal discovery algorithms to find reliable causal structure from limited client data.

Federated learning (FL), as a secure framework for cooperative training with multiple clients, learns a unified model from scattered data by exchanging model parameters or gradients among clients, without exposing local clients' data \cite{mcmahan2017communication}. FL has made good progress in areas such as image classification  and recommendation systems \cite{chai2020secure}. However, recent FL algorithms are based on continuous optimization and cannot be directly applied to causal discovery algorithms with a combinatorial optimization property. For example, constraint-based algorithms PC \cite{spirtes2000causation} and FCI \cite{spirtes2013causal} use conditional independence between variables to judge whether there is a causal structure, while score-based algorithms GES \cite{chickering2002optimal} and the max-min hill-climbing algorithm \cite{tsamardinos2006max} use score functions and heuristic search strategies to find causal graphs with the best scores.

Recent gradient-based causal discovery algorithms, \citet{zheng2018dags} and \citet{yu2019dag} make use of smooth equality constraints instead of discrete acyclic constraints to discover causal structures through continuous optimization (i.e., Augmented Lagrange method \cite{nemirovsky1999optimization}), and they provide the opportunity of learning  causal graphs in a continuous manner within a federated framework. Subsequently, \citet{lachapelle2019gradient} and \citet{zheng2020learning} introduced deep neural networks to deal with more complex causal models. However, these gradient-based algorithms need to center data, which is not feasible due to privacy protection. For dispersed data, several distributed causal discovery methods have been proposed. Some of them need to make assumptions about parameters but lack  generalization \cite{shimizu2012joint,xiong2021federated}, and  others need to share additional learning parameters but cause privacy leakage \cite{na2010distributed,ye2022distributed}. More importantly, they are often unable to deal with non-independent identical distributed (non-iid) data among clients.

\begin{figure}[ht]
\centering
\subfigbottomskip=2pt
\includegraphics[scale=0.69]{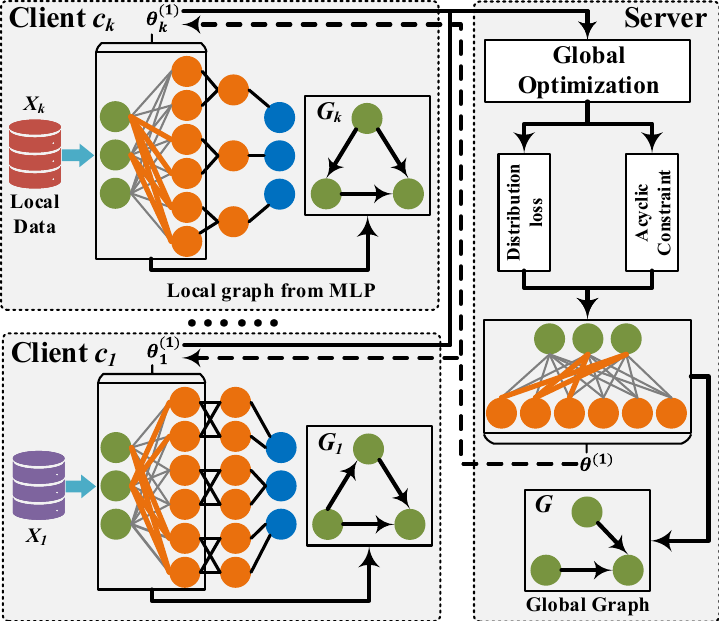} 
\caption{Schematic framework of FedCausal on the non-parametric model. In each interaction, the clients optimize local causal models based on local data and send the first layer parameters $\theta_k^{(1)}$ to the server. The server optimizes the global model to conform to the local distributions and explicitly constrains its acyclicity to obtain a global causal graph $G$. FedCausal then broadcasts server parameters $\theta^{(1)}$ to clients for the next round interaction.}
\label{fig1}
\end{figure}

In this paper, we develop a general federated DAG learning strategy (FedCausal) to seek the global causal graph from horizontally partitioned data with different distributions. This strategy introduces the centralized DAG learning framework into federated training by adding a proximal term, and designs a global optimization process instead of traditional weighted average in the server to aggregate local causal graphs, which naturally ensures the sparsity and acyclicity of the global causal graph. The local and global optimization processes form a whole as an explainable adaptive optimization objective, which is consistent with the centralized optimization objective under statistical homogeneity. In addition, this explainable objective allows clients to flexibly learn local data distributions with statistical heterogeneity and seek a uniform global graph on the server. Figure \ref{fig1} shows the conceptual framework of FedCausal. Our main contributions can be outlined as follows:\\
\noindent(i) We meticulous extend the centralized DAG learning framework to federated scenarios. FedCausal explicitly constrains the acyclicity of the global causal graph and optimizes it to conform to  dispersed data, ensuring effective causal information interaction between local and global.\\
\noindent(ii) FedCausal unifies the global aggregation optimization and local DAG optimization, and formulates a complete causal graph learning process. We prove that its optimization process is consistent with centralized DAG optimization on homogeneous data and is suitable for heterogeneous data.\\
\noindent(iii) We conduct experiments on synthetic and real datasets, and prove that FedCausal is close to or even better than  centralized learning on homogeneous data, and outputs more accurate and acyclic causal graphs than other methods \cite{ng2022towards,gao2023feddag} on heterogeneous data.

\section{Related Work}
The number of DAGs is super-exponential in the number of variables. As such, learning discrete DAGs is generally NP-hard for traditional causal discovery algorithms \cite{spirtes2000causation,chickering2002optimal,bernstein2020ordering}. To avoid the difficult combinatorial optimization of the acyclic constraint, NOTEARS \cite{zheng2018dags} transforms the acyclic constraint into a smooth equality constraint, and converts the causal discovery problem into a continuous optimization one with efficient solvers. DAG-GNN \cite{yu2019dag} extends NOTEARS using graph neural networks to handle various variable types. Gran-DAG \cite{lachapelle2019gradient} uses neural networks to deal with non-linear causal relationships. NOTEARS-MLP \cite{zheng2020learning} proposes a generalized function model and characterizes a non-parametric structure equation model of DAG via partial derivatives. HetDAG \cite{liang2023directed} extends NOTEARS-MLP to learn the DAGs of attributed heterogeneous network, and DARING \cite{he2021daring} considers the independence of heterogeneous noises to improve the robustness. MCSL \cite{ng2022masked} adopts a $\mathrm{Gumbel-Sigmoid}$ function to approximate the parameter matrix of a DAG into a binary adjacency one. However, the above approaches target at centralized data, and their assumptions or strategies for identifying causal graphs cannot cope with statistically heterogeneous data. Based on the continuous optimization framework, we design a global optimization to flexibly learn the global causal graph by aggregating local causal graphs from decentralized data that may have different distributions.

Several attempts have been made to learn the causal structure from distributed datasets with different subsets of variables. \cite{danks2008integrating} estimated the local partial ancestral graph from each dataset, and then found global partial ancestral graphs on the complete set of variables with the same $d$-connection and $d$-separation relationships as the local PAGs. \cite{triantafillou2015constraint} used a SAT solver in a similar process to improve the algorithm's scalability. However, these methods cannot uniquely identify the causal graph and suffer from a large indeterminacy. For the horizontally distributed dataset, \citet{gou2007learning} and \citet{na2010distributed} used a two-step process that independently estimates the Bayesian network structure by each client's local dataset, then performs conditional independent testing and voting to obtain the final causal graph. These distributed methods separately use local datasets during training and directly share local models parameters, which result in poor performance and privacy leakage. Our FedCausal aggregates the global causal graph using a very small fraction of the parameters of local models, effectively avoiding the risk of reconstructing local data from model parameters. It does not simply aggregate the local causal graphs, but adds a proximal term and global optimization to enable sufficient information to be exchanged between clients and the server.

The recent NOTEARS-ADMM approach \cite{ng2022towards} adopts the alternating direction method of multipliers to decompose the continuous optimization objective of gradient-based methods, and learns the global causal structure by exchanging client model parameters. However, NOTEARS-ADMM requires all clients to participate in the training and makes all local models consistent, and thus lacks flexibility and does not account for statistical heterogeneity. Although FedDAG \cite{gao2023feddag} takes into account  heterogeneous data, it directly masks local models with a global binary matrix. As a consequence, it cannot effectively update the masked model parameters and obtain accurate causal graphs. Both NOTEARS-ADMM and FedDAG enforce too stringent local models and suffer from performance degradation. FedCausal naturally combines local and global optimizations, forms an adaptive optimization process, and has a consistent optimization objective with the centralized DAG learning under iid data without additional assumptions. Compared with existing federated DAG learning methods, FedCausal is more interpretable and more suitable for non-iid data, it trains local models without strong restrictions to learn the unified global graph. FedCausal accurately identifies the unique causal graph while conforming to the heterogeneous distribution of local data.

\section{Background}
The causal discovery problem can be defined as follows: given the observed data $X\in \mathbb{R}^{n\times d}$ and the variables' set $V=\{V_1,\cdots,V_d\}$, we need to learn a DAG $\mathcal{G}$ for the joint distribution $P(X)$. In the federated scenario, this problem becomes more complicated: let $\mathcal{C}=\{c_1,\cdots,c_k\}$ represent the clients' set, $X=\{X_1,\cdots,X_k\}$ represent the local clients' data, and $N=\{n_1,\cdots,n_k\}$ represent the sample size of clients' data; we need to learn a uniform causal graph $\mathcal{G}$ without exposing clients' data. The difficulty is that clients' data may be non-iid. In other words, let $P=\{P_1,\cdots,P_k\}$ be the joint distributions of clients' data. There are at least two clients $c_i$ and $c_j$ with $P_i\neq P_j$. Even though existing DAG learning methods based on continuous optimization  mostly focus on centralized data and cannot handle statistical heterogeneity, they are necessary for federated causal discovery because most of the federated learning algorithms adopt continuous optimization. For this reason we introduce next some of the existing continuous DAG learning strategies.

\subsection{Continuous optimization for DAG learning}
NOTEARS uses equality to constrain the acyclicity of causal graphs and continuously optimizes a causal graph on the linear data model. NOTEARS defines a weight matrix $W$ to represent the causal model, thus $W_{ij}=0$ if and only if there is no edge from $V_i$ to $V_j$ in the real causal graph $\mathcal{G}$. Then, it uses the equation $h(W)=tr(e^{W\circ W})-d=0$ to replace the discrete constraint $\mathcal{G}\in DAGs$ (see \cite{zheng2018dags} for details). With this constraint, NOTEARS reconstructs data from $W$ to minimize the reconstruction residual and designs the following constrained optimization formula:
\begin{gather}
        \min_{W\in\mathbb{R}^{d\times d}}L(W;X)=\frac{1}{2n}||X-XW||_2^2+\lambda||W||_1 \nonumber\\
        \text{s.t. }h(W)=tr(e^{W\circ W})-d=0
    \label{eq1}
\end{gather}%
where $||W||_1$ guarantees the sparsity of the causal graph, and $\lambda$ is the hyper-parameter. The approximate solution is sought via the L-BFGS-B algorithm \cite{byrd1995limited}.

To extend NOTEARS to non-parametric models, \citet{zheng2020learning} proposed NOTEARS-MLP, which learns the causal model via a multi-layer perceptron (MLP) and uses partial derivatives to express the dependence between variables. NOTEARS-MLP learns the causal generation model $f=\{f_1,\cdots,f_d\}$ for each variable. It is easy to show that there is no edge from $V_j$ to $V_i$, if and only if the partial derivative $\partial_jf_i$ of $f_i$ with respect to $X_j$ is equal to $0$. So, NOTEARS-MLP uses the partial derivatives of $f$ to represent the parameter matrix $W(f)$: $[W(f)]_{ji}=||\partial_jf_i||_2$. Let $\theta=\{(A_1^{(1)},\cdots,A_1^{(h)}),\cdots,(A_d^{(1)},\cdots,A_d^{(h)})\}$ be the parameters of all MLPs and $A_i^{(h)}$ be the $h$-th layer parameter of the MLP corresponding to $f_i$, NOTEARS-MLP uses the first layer parameters $\theta^{(1)}=\{A_1^{(1)},\cdots,A_d^{(1)}\}$ of all MLPs to represent the weight matrix $[W(\theta^{(1)})]_{ji}=||A_{i,(:,j)}^{(1)}||_2$. The constrained optimization formula of NOTEARS-MLP is as follows:
\begin{gather}
    \min_{\theta}L(\theta;X)=\frac{1}{2n}[||X-\mathrm{mlp}(X;\theta)||_2^2+\lambda||\theta||_1] \nonumber\\
    \text{s.t. }h(\theta^{(1)})=tr(e^{W(\theta^{(1)})\circ W(\theta^{(1)})})-d=0
    \label{eq2}
\end{gather}%
where $\mathrm{mlp}(X;\theta)$ is the reconstruction result of original data $X$ from all causal generation models parameterized by MLP and $||\theta||_1$ guarantees the  causal model sparsity. 

Both NOTEARS and NOTEARS-MLP require centerized iid data. In contrast, FedCausal targets decentralized non-iid data. It unifies local DAG learning on non-iid data into a global optimization framework, which improves the causality learning capability of client models under limited local data, and ensures unity and accuracy of global causal graphs under statistical heterogeneity. The global model of FedCausal corresponds to the unique causal graph that explains causality between variables, and the local models learn the heterogeneous distributions of clients' data.

\section{Proposed Methodology}
\label{section 4}

\subsection{Optimization for linear SEM}
To learn causal graphs that conform to the data distribution, both centralized and federated DAG learning methods aim to minimize the reconstruction residual of observational data and guarantee the graph acyclicity. For a linear structural equation model (SEM), assume that there are $K$ clients holding $n=\sum^K_{k=1}n_k$ samples in total. The global weight matrix $W$ can be learned as follows: 
\begin{gather}
        W = \arg\min_{W}\sum^K_{k=1} \frac{1}{2n_k} [||X_k-X_kW||_2^2+\lambda||W||_1] \nonumber\\ 
        \text{s.t. }tr(e^{W\circ W})-d=0
    \label{eq3}
\end{gather}%
However, clients cannot expose their own data and they can only use local data to optimize the local weight matrix $W_k$. In addition, the acyclic constraint term is applied to the local matrix $W_k$ rather than the global $W$. A natural idea is to aggregate the local matrices into a global one with guaranteed sparsity and acyclicity.

Without loss of generality, we assume that the probability of the observational data being distributed over client $c_k$ is equal to the frequency $n_k/n$ of samples appearing in $c_k$, given sufficient data. In other words, observational data match the joint distribution of client $c_k$ with probability $n_k/n$. To measure local empirical risks over possible data distributions, the global matrix should be the weighted average of the local matrices based on sample sizes: $W=\sum^K_{k=1}\frac{n_k}{n}W_k$. Unfortunately, clients may learn weight matrices with opposite causal relationships, which introduces a strong numerical bias to the global matrix and violates the acyclicity, and thus fails to reconstruct original data if we optimize its acyclicity without data participation. To solve this problem, we design a global constrained optimization formula to aggregate local matrices and ensure the global matrices' sparsity and acyclicity:
\begin{gather}
        \min_{W}L(W)=\sum^K_{k=1}[\frac{n_k}{n}||W-W_k||_2^2] \nonumber\\ 
        \text{s.t. }h(W)=tr(e^{W\circ W})-d=0
    \label{eq4}
\end{gather}%
where $n_k/n||W-W_k||_2^2$ is the distribution loss term to constrain the global matrix to approach the local matrix. Obviously, Eq. \eqref{eq4} measures and minimizes local empirical risks, and its equality constraint also explicitly constrains the acyclicity of the global causal graph. Since the global matrix resulting from aggregating sparse local matrices is also numerically sparse, we do not add the sparse penalty term to Eq. \eqref{eq4}. We broadcast the learned global matrix $W$ to all clients and further optimize their local matrices as follows:
\begin{gather}
        \min_{W_k}L_k(W_k;X_k)=\frac{1}{2n_k}||X_k-X_kW_k||_2^2+\lambda_1||W_k||_1 \nonumber\\ 
        \text{s.t. }h(W_k)=tr(e^{W_k\circ W_k})-d=0
    \label{eq5}
\end{gather}%
Local matrix optimization still requires the sparsity penalty and acyclic constraint to prevent its overfitting. In practice, we find that adding a penalty term $\lambda_2||W_k-W||_2^2$ to bring local matrices close to the global one is beneficial for the convergence of local optimization. This penalty term is weaker than the equality constraint of NOTEARS-ADMM \cite{ng2022towards}, which forces the local matrix to be equal to the global matrix, but it is more flexible and suitable for addressing the statistical heterogeneity. FedCausal first optimizes local matrices $\{W_k\}_{k=1}^K$ at clients using Eq. \eqref{eq5} and uploads them to the server. The server then optimizes the global matrix $W$ using Eq. \eqref{eq4} and distributes it to the local models. FedCausal repeats the above local and global optimizations for a sufficiently large number of iterations, or stops when the acyclic constraint term falls below a pre-defined threshold.

\subsection{Optimization for non-parametric SEM}

The non-parametric model is more complex than the linear model, since it has more parameters and more diverse clients' data distributions. Exchanging all models' parameters between the clients and the server not only causes excessive communication overheads, but also leads to the same local models and to the violation of statistical heterogeneity. To address these issues, we adjust the global optimization and pass only a few parameters between the clients and the server. We denote the parameters of the local causal generation model as $\theta_k$, the parameters of the first layer as $\theta_k^{(1)}$ and those of other layers as $\theta_k^{(-1)}$. The global optimization formula of FedCausal in the non-parametric model is:
\begin{gather}
    \min_{\theta^{(1)}}L(\theta^{(1)})=\sum^K_{k=1}[\frac{n_k}{n}||\theta^{(1)}-\theta_k^{(1)}||_2^2] \nonumber\\
    \text{s.t. }h(\theta^{(1)})=tr(e^{W(\theta^{(1)})\circ W(\theta^{(1)})})-d=0
    \label{eq6}
\end{gather}%
FedCausal receives only the first layer parameters $\theta_k^{(1)}$ of the local models to optimize the global model. The unified global causal graph is derived from the partial derivative of the first-layer global model parameters $\theta^{(1)}$, so there is no need to send $\theta_k$. This design not only reduces the communication overhead compared to passing all parameters as in NOTEARS-ADMM \cite{ng2022towards}, but also enforces privacy protection, because it is difficult for attackers to reconstruct data from the first layer parameters of a complex model. We then  broadcast $\theta^{(1)}$ to all clients, replace their $\theta_k^{(1)}$, and optimize their $\theta_k$ with the following formula:
\begin{gather}
    \min_{\theta_k}L_k(\theta_k;X_k)=\frac{1}{2n_k}||X_k-\mathrm{mlp}(X_k;\theta_k)||_2^2+\lambda_1||\theta_k^{(1)}||_1 \nonumber\\
    \text{s.t. }h(\theta_k^{(1)})=tr(e^{W(\theta_k^{(1)})\circ W(\theta_k^{(1)})})-d=0
    \label{eq7}
\end{gather}%
%Eq. \eqref{eq7} can also add a penalty term $\lambda_2||\theta_k^{(1)}-\theta^{(1)}||_2^2$ to curate local graphs close to the global graph. 
FedCausal learns the global causal graph using Eq. \eqref{eq6}, allows clients to learn different causal models conforming to non-iid local data using Eq. \eqref{eq7}, and thus provides high flexibility for causal discovery in non-iid data. Algorithm \ref{algorithm1} gives the main process of FedCausal on non-parametric data, where $\alpha$ and $\rho$ are the Lagrangian multiplier and the penalty parameter, $h_{tol}$ and $\rho_{max}$ are the thresholds that control the loop termination, $\gamma$ and $\beta$ are the condition and the step size that control the update of $\rho$ and $\alpha$. Detailed parameter settings are included in the supplementary file.

\begin{algorithm}[htbp]
\caption{\textbf{FedCausal}: Federated Causal Learning }
\label{algorithm1}
\textbf{Input}: Observed data $X=\{X_1,\cdots, X_K\}$\\
\textbf{Output}: Causal weight matrix $W(\theta^{(1)})$\\
\begin{algorithmic}[1]
\STATE Initial global parameters $\theta^{(1)}$, local parameters $\{\theta_k=(\theta^{(1)}_k,\theta^{(-1)}_k)\}^K_{k=1}$ and hyperparameter-list $\{\alpha,\rho,h_{tol},\rho_{max},\gamma,\beta\}$
\FOR{$t=0,1,2,\cdots$}
\WHILE{$\rho<\rho_{max}$}
\STATE Broadcast $\theta^{(1)},\alpha,\rho$: $\theta^{(1)}_k=\theta^{(1)},\alpha_k=\alpha,\rho_k=\rho$
\STATE Update local parameters $\{\theta_k\}^K_{k=1}$ using Eq. \eqref{eq7}
\STATE Upload $\{\theta^{(1)}_k\}^K_{k=1}$ to the server
\STATE Update global parameters $\theta^{(1)}$ using Eq. \eqref{eq6}
\IF{$h(\theta^{(1)})>\gamma H$}
\STATE $\rho=\beta\rho$
\ELSE
\STATE Break
\ENDIF
\ENDWHILE
\STATE Set $H=h(\theta^{(1)})$, $\alpha=\alpha+\rho h(\theta^{(1)})$
\IF{$h(\theta^{(1)})<=h_{tol}$ or $\rho>=\rho_{max}$}
\STATE Break
\ENDIF
\ENDFOR
\STATE \textbf{return} $W(\theta^{(1)})$
\end{algorithmic}
\end{algorithm}

Let's consider the case of $K$ client models with one hidden layer of $m$ units. FedCausal's clients take $O(n_kd^2m+d^2m+d^3)$ to compute the objective and the gradient. The computation complexity of FedDAG's clients is $O(n_kd^2m+n_kd^2+d^2m+d^3)$, due to the additional mask operation. NOTEARS-ADMM does not compute the acyclic term on clients, so its clients' computation complexity is $O(n_kd^2m+d^2m)$, but it suffers from overfitting local models. Both FedCausal and NOTEARS-ADMM optimize the global graph on the server, so their complexity on the server side is $O(Kd^2m+d^3)$. However, FedCausal only needs to compute the first layer parameters of the model, so it is usually faster than NOTEARS-ADMM, especially when the model has at least two layers. The communication overheads of FedCausal, NOTEARS-ADMM, and FedDAG for a single round interaction are $d^2m$, $d^2m+dm$ and $d^2$, respectively. As the model becomes complex with more layers, the  overhead of NOTEARS-ADMM sharply increases. FedDAG  averages only local matrices and has a lower server computation than NOTEARS-ADMM and FedCausal, but its global matrix may not uncover an accurate global causal graph, as we  show in our experiments.

\subsection{Analysis of Explainable Adaptive Optimization}
\newtheorem{lemma}{Lemma}
{\begin{lemma}Under the condition that local data are homogeneous and obey \textbf{linear SEM}, the optimization objective of FedCausal is consistent with that of centralized DAG learning, i.e. the global matrix $W\rightarrow\mathop{\arg\min}_W||X-XW||_2^2/2n+\lambda||W||_1+\rho h^2(W)/2+\alpha h(W)$.\end{lemma}}

For the linear data model, the optimization objective of the penalty term $||W-W_k||_2^2$ in Eq. \eqref{eq4} is $W\rightarrow W_k$, and that of Eq. \eqref{eq5} is $W_k\rightarrow\mathop{\arg\min}_{W_k}L_k(W_k;X_k)+\rho_kh^2(W_k)/2+\alpha_kh(W_k)$, so the actual optimization objective of $||W-W_k||_2^2$ is $W\rightarrow\mathop{\arg\min}_{W}L_k(W;X_k)+\rho_kh^2(W)/2+\alpha_kh(W)$. We can then replace the penalty term in the global optimization (Eq. \eqref{eq4}) with the actual objective, and obtain the following:
\begin{gather}
        W\rightarrow\mathop{\arg\min}_W\sum^K_{k=1}\frac{n_k}{n}[\frac{1}{2n_k}||X_k-X_kW||_2^2+\lambda||W||_1] \nonumber\\
        +\frac{\rho}{2}h^2(W)+\alpha h(W)
    \label{eq8}
\end{gather}%
{Eq. \eqref{eq8} only preserves the global acyclic constraint term, because the local acyclic constraint acts on the local matrices and does not constrain the global matrix.} By merging the terms in Eq. \eqref{eq8}, we find that the optimization objective of the global matrix is: $W\rightarrow\mathop{\arg\min}_W||X-XW||_2^2/2n+\lambda||W||_1+\rho h^2(W)/2+\alpha h(W)$. Obviously, this optimization objective is consistent with the centralized optimization objective of NOTEARS, which means that FedCausal may achieve the same results as NOTEARS, even though the data are scattered.

\begin{lemma}Under the condition that local data are homogeneous and obey \textbf{non-parametric SEM}, the optimization objective of FedCausal is consistent with that of centralized DAG learning, i.e. the global parameters $\theta^{(1)}\rightarrow\mathop{\arg\min}_{\theta^{(1)}}||X-\mathrm{mlp}(X;\theta^{(1)},\theta^{(-1)})||_2^2/2n+\lambda||\theta^{(1)}||_1+\rho h^2(\theta^{(1)})/2+\alpha h(\theta^{(1)})$.\end{lemma}

For the non-parametric data model, the optimization objective of the penalty term $||\theta^{(1)}-\theta^{(1)}_k||_2^2$ in Eq. \eqref{eq7} is $\theta^{(1)}\rightarrow\theta^{(1)}_k$. By expressing the local parameters $\theta_k$ optimized by Eq. \eqref{eq6} as $(\theta^{(1)}_k,\theta^{(-1)}_k)$, we obtain the actual optimization objective of $||\theta^{(1)}-\theta^{(1)}_k||_2^2$ as $\theta^{(1)}\rightarrow\mathop{\arg\min}_{\theta^{(1)}}L_k(\theta^{(1)},\theta^{(-1)}_k;X_k)+\rho_kh^2(\theta^{(1)})/2+\alpha_kh(\theta^{(1)})$. Similarly, we can deduce the global optimization objective in the non-parametric data model as:
\begin{gather}
        \theta^{(1)}\rightarrow\mathop{\arg\min}_{\theta^{(1)}}\sum^K_{k=1}\frac{n_k}{n}[\frac{1}{2n_k}||X_k-\mathrm{mlp}(X_k;\theta^{(1)},\theta_k^{(-1)})||_2^2 \nonumber \\
        +\lambda||\theta^{(1)}||_1]+\frac{\rho}{2}h^2(\theta^{(1)})+\alpha h(\theta^{(1)})
    \label{eq9}
\end{gather}%
We consider two cases. On the one hand, if clients' data are \emph{homogeneous}, the local parameter $\theta_k^{(-1)}$ can be the same or very similar, so we can merge the terms in Eq. \eqref{eq9} to obtain the final optimization objective: $\theta^{(1)}\rightarrow\mathop{\arg\min}_{\theta^{(1)}}||X-\mathrm{mlp}(X;\theta^{(1)},\theta^{(-1)})||_2^2/2n+\lambda||\theta^{(1)}||_1+\rho h^2(\theta^{(1)})/2+\alpha h(\theta^{(1)})$. This optimization objective is also consistent with the centralized optimization objective of NOTEARS-MLP. On the other hand, if clients have decentralized \emph{heterogeneous} data, Eq. \eqref{eq9} allows clients to learn different $\theta_k^{(-1)}$, and still optimizes a unified $\theta^{(1)}$. In other words, regardless whether the local data fit linear or nonlinear models, FedCausal learns different data joint distributions on the clients and optimizes a uniform causal graph on the server. In other words, FedCausal adapts to the case where clients have {different causal generation models but the same causal graph}. In summary,  our FedCausal can flexibly adapt to statistical heterogeneous/homogeneous data with an explainable proof.

\section{Experiments}
We compare FedCausal against  recent federated causal discovery algorithms, including FedDAG \cite{gao2023feddag} and NOTEARS-ADMM (abbreviated as NO-ADMM) \cite{ng2022towards}. We also set three baselines based on  NOTEARS \cite{zheng2018dags}. The first baseline (NO-ALL) learns DAGs from all data by centering local data; the second (NO-Avg) independently learns local causal graphs and combines these graphs by weighted average; the third one (NO-w/oAcy) is similar to our strategy but disregards the acyclic constraint on the global causal graph. For nonlinear or heterogeneous data, we adapt the nonlinear or non-parametric version of FedCausal, FedDAG and NO-ADMM, and replace NOTEARS with NOTEARS-MLP \cite{zheng2020learning} for NO-ALL, NO-Avg and NO-w/oAcy. For real datasets, we add two classic baselines PC \cite{spirtes2000causation} and GES \cite{chickering2002optimal}, which use all data but identify complete partially DAG (CPDAG) with undirected edges. We provide the hyperparameter settings of FedCausal and other baselines in the supplementary file.

We study the performance of the compared methods on synthetic iid and non-iid datasets, as well as real datasets.  We use the Erdös–Rényi (ER) and Scale-free (SF) models to generate the ground truth DAGs. We use the linear Gaussian (LG) model and the additive noise model with MLP (ANM-MLP) \cite{zheng2020learning} respectively to synthesize linear and nonlinear data. For non-iid data, we fix the generated DAG and randomly sample models from the linear Gaussian model (LG), the additive noise model with MLP (ANM-MLP), the additive model with Gaussian processes (ADD-GP) \cite{buhlmann2014cam} and the additive index model (MIM) \cite{yuan2011identifiability} to generate data for different clients. We guarantee that, even if different clients choose the same model, their model parameters will be different. We use the same server (Ubuntu 18.04.5, Intel Xeon Gold 6248R and Nvidia RTX 3090) to perform experiments and report the structural hamming distance (SHD), true positive rate (TPR) and false discovery rate (FDR) of the estimated DAGs, averaged over 10 random runs. A larger TPR indicates a better performance, while the opposite trend holds for SHD and FDR. 

\subsection{Results on iid data}
\begin{figure}[ht]
\centering
\subfigure[Results on linear model]{
\includegraphics[scale=0.16]{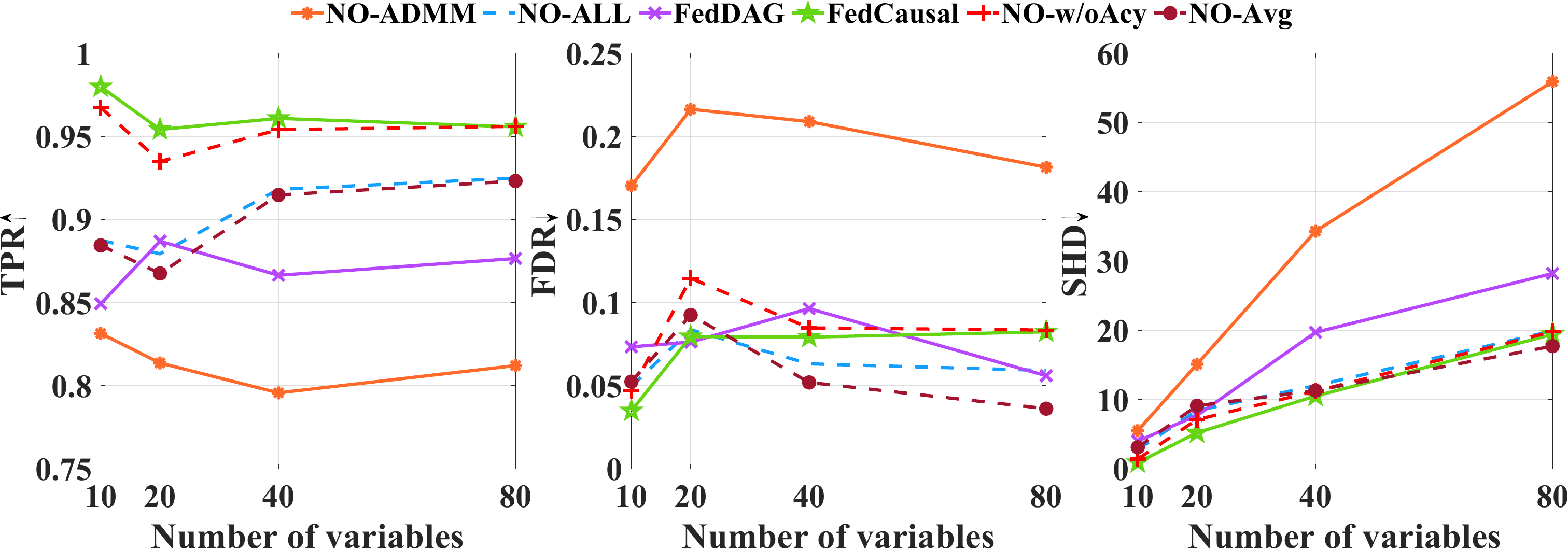}
}
\subfigure[Results on nonlinear model]{
\includegraphics[scale=0.16]{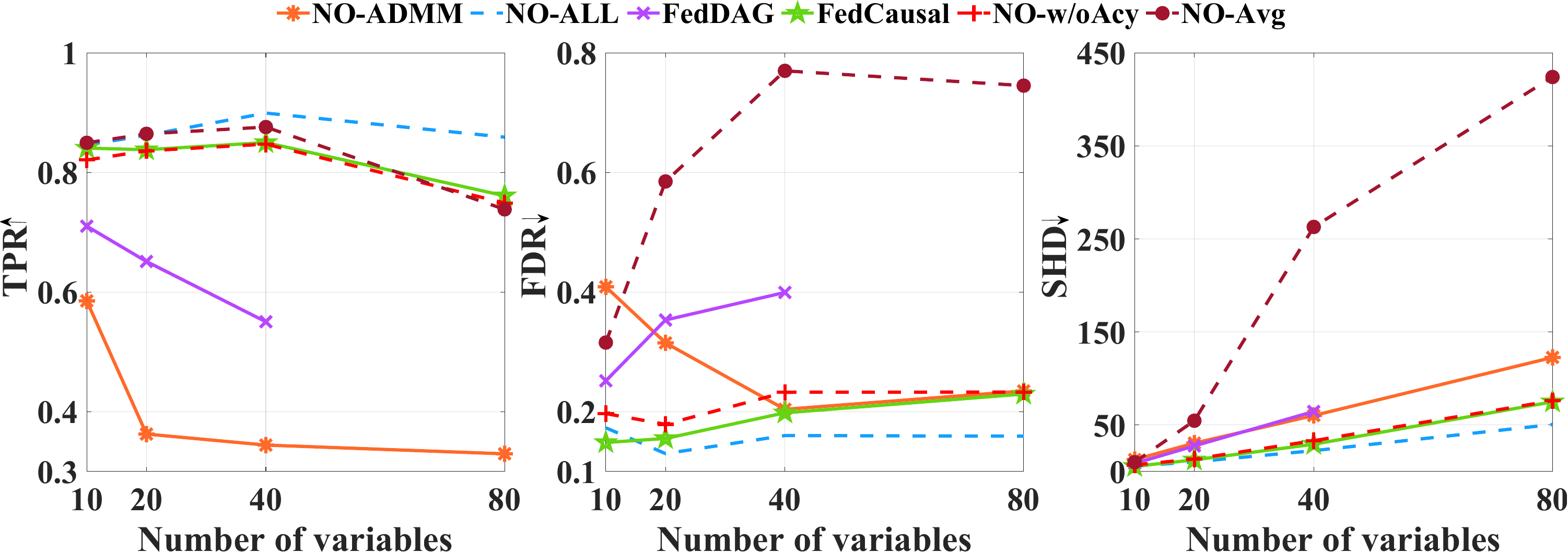}
}
\caption{Results on linear and non-linear iid data. Solid lines correspond to federated methods and dotted lines are NOTEARS-based baselines.}
\label{fig2}
\end{figure}

To test FedCausal on iid data, we set up 10 clients and generate 200 samples for each client with linear (LG) and nonlinear (ANM-MLP) iid data models. We conduct the experiment to estimate randomly generated DAGs with $\{10,20,40,80\}$ variables. Based on the results of FedCausal and other baselines in Figure \ref{fig2}, we have some observations:

\noindent (i) FedCausal obtains the highest TPR on the linear model across different numbers of variables. NO-ALL learns the causal graph using all data, so its results can be taken as an upper bound.  FedCausal even outperforms NO-ALL, probably because there are sufficient data for clients to learn local causal graphs on the simple linear model. In addition, the collaboration between clients works alike an ensemble system that helps to reinforce correct edges and remove erroneous ones. NO-ADMM achieves the lowest TPR and highest FDR and SHD, since it does not consider the sparsity and acyclicity on clients, which cause overfitted local graphs and fail to form a good global causal graph. FedDAG performs well in the linear model, approaching NO-ALL with only a few variables; it does not with more variables. NO-Avg performs almost as well as NO-ALL on the linear model, since the linear model is simple and clients can confidently learn local graphs without any collaboration.

\noindent (ii) FedCausal also achieves the lowest FDR and SHD values on the nonlinear model, being clearly superior to NO-ALL, with very close TPR values. This is because our global optimization aligns with the centralized optimization objective of NO-ALL that uses all data, and federated learning allows clients to collaborate with each other to reduce erroneous edges. For the same reason as for the linear model, NO-ADMM still performs poorly on the nonlinear model. FedDAG does not perform as well on the nonlinear model, as it did on the linear one, because the \texttt{Gumbel-Sigmoid} function it uses is not suitable for dealing with nonlinear causality, which leads to biased results. We do not report the results of FedDAG on 80 nodes, since \citet{gao2023feddag} only provided the hyperparameters of FedDAG on 10, 20, and 40 nodes, and FedDAG is very sensitive to the hyperparameters and fails on 80 nodes. NO-Avg yields unreliable results with very high FDR and SHD values, even though its TPR is also high. This indicates that simply averaging local causal graphs may result in more erroneous edges.

\noindent (iii) Our aggregation optimization effectively improves the performance of FedCausal. Compared to NO-w/oAcy, FedCausal performs better on the linear model and obtains significantly lower FDR and SHD values on the nonlinear model. This is because our adaptive optimization not only measures the local empirical risk, but also constrains the acyclicity of the global graph, while NO-w/oAcy disregards this constraint. FedCausal trains the global graph to conform to the local data distribution and eliminates the numerical bias caused by aggregation that may lead to a cyclic graph, thus effectively improves the global graph.

\subsection{Results on non-iid data}
In this experiment, we also set 10 clients, 200 samples for each client and graphs with $\{10, 20, 40, 80\}$ variables. We randomly select models from LG, ANM-MLP, ADD-GP or MiM to generate non-iid data for different clients. Figure \ref{fig3} shows the results of FedCausal and other baselines. We have some important observations:

\begin{figure}[ht]
\centering
\subfigure[Results on ER graphs]{
\includegraphics[width=8cm]{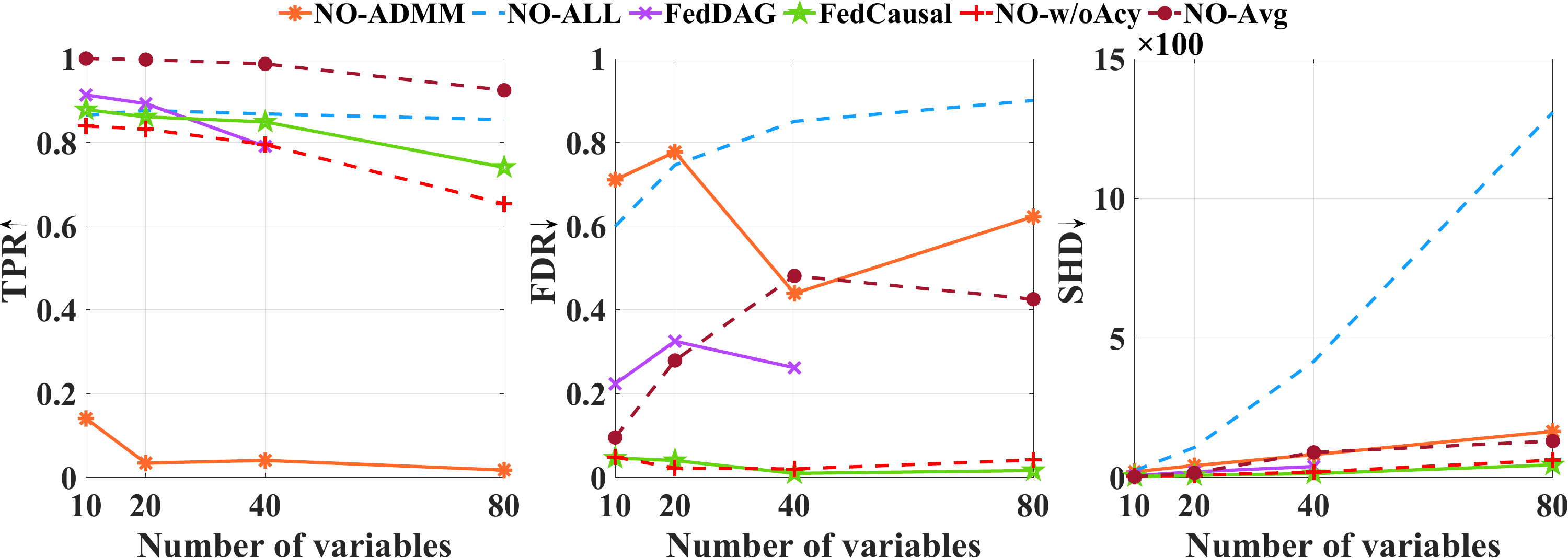}
}
\subfigure[Results on SF graphs]{
\includegraphics[scale=0.16]{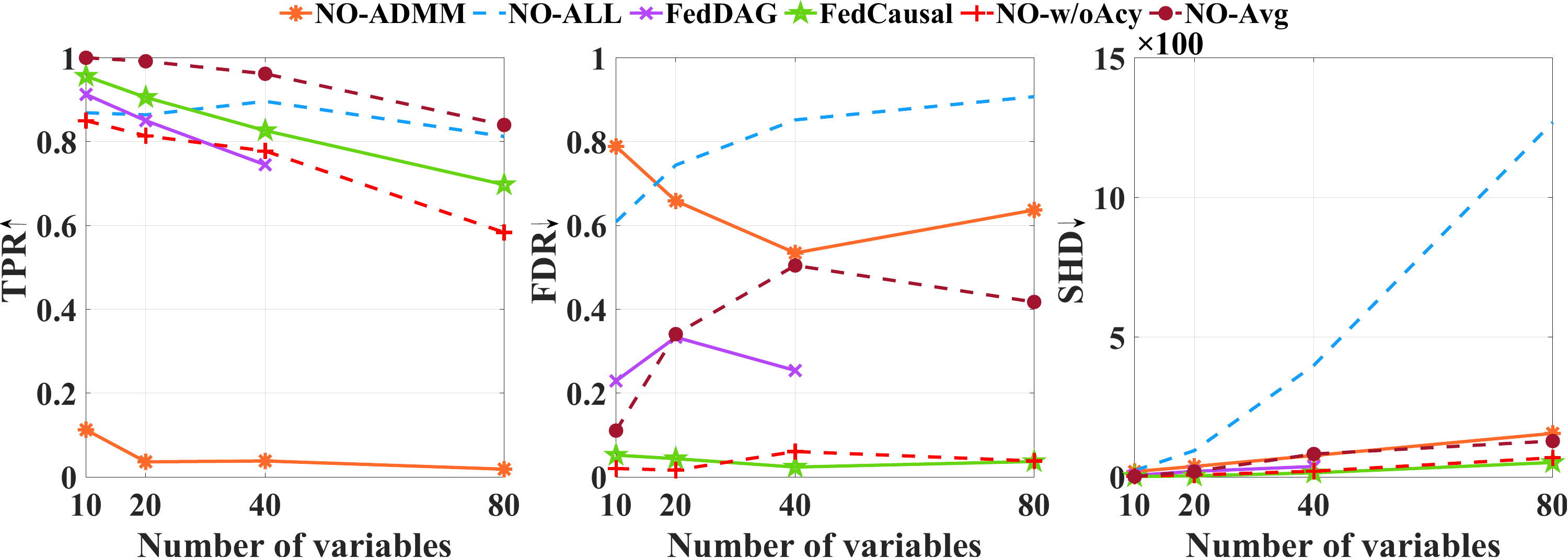}
}
\caption{Results on linear and non-linear non-iid data.}
\label{fig3}
\end{figure}

\noindent (i) FedCausal manifests the best performance among the compared methods on decentralized heterogeneous data. NO-ADMM pursues the consistency of all local models and thus does not identify the global causal graph in this practical and challenging non-iid setting. NO-ALL gets extremely high FDR values because the training data are mixed with multiple causal models, which prevent NO-ALL to accurately identify causal relationships. This observation suggests that causal algorithms with the prerequisite of centralized data cannot identify the causal graphs from non-iid data. Federated causal discovery algorithms (FedCausal, FedDAG, and NO-w/oAcy) perform better on heterogeneous data than on homogeneous nonlinear data. This is because clients discover the same false causal relationships on iid data and reinforce these false edges in aggregation, while clients on non-iid data can offset the false edges. FedCausal has a much lower FDR than FedDAG and a higher TPR than NO-w/oAcy. The graph found by NO-Avg has a very high TPR and FDR values without guaranteed acyclicity.

\noindent (ii) FedCausal effectively improves the local causal graphs and guarantees the acyclicity. NO-Avg cannot obtain a reliable and acyclic causal graph on non-iid data, due to the poor local results and high $h(\theta^{(1)})$ (a smaller $h(\theta^{(1)})$ indicates a better acyclicity) in Table \ref{tabel2}. This is because the clients of NO-Avg are trained independently, but the limited local data can't support them to accurately learn the nonlinear and heterogeneous data models, which are more complex than linear ones. In contrast, by aggregating local results at each interaction, FedCausal, NO-w/oAcy and FedDAG  significantly improve the quality of local graphs. FedDAG learns local graphs masked by an approximate binary global matrix, which allows clients to get very similar results with global $h(\theta^{(1)})$ equal to 0. {However, the masked model parameters cannot be effectively updated, which causes the model to be unable to correct error edges and limits its accuracy.} In terms of the client performance and global acyclicity, FedCausal is superior to NO-w/oAcy. This proves the effectiveness of our adaptive optimization that guarantees the acyclic global causal graph for clients.

In addition, to study the communication efficiency of FedCausal and other federated baselines, we show the performance trend of the algorithm as a function of the number of communications in the supplementary file. The results show that FedCausal has the most superior communication efficiency. We also explore the robustness of FedCausal against different numbers of clients and local sample sizes, and present the results in the supplementary file. We observe that FedCausal is stable and in all cases outperforms the rivals.

\begin{table}[htbp]
  \centering
  \begin{tabular}{l| r r| l}
    \hline
    \multirow{2}{*}{40 nodes}   &\multicolumn{2}{c|}{Metrics of local graphs}   &\multirow{2}{*}{$h(\theta^{(1)})\downarrow$}\\
    \cline{2-3}
            &TPR$\uparrow$      &FDR$\downarrow$    &  \\
    \hline
    FedDAG  &82.4$\pm$0.0       &10.8$\pm$0.3       &\bf{$\approx$0} \\
    FedCausal&\bf{92.4$\pm$1.9}  &\bf{4.4$\pm$2.8}   &\bf{$\approx$4.2$\times\bf{10^{-12}}$}\\
    NO-w/oAcy&84.7$\pm$3.4       &\bf{8.3$\pm$5.0}   &$\approx$7.8$\times10^{-7}$\\
    NO-Avg  &77.0$\pm$11.2      &71.9$\pm$6.0       &$\approx$2.2$\times10^{-1}$\\
    \hline
  \end{tabular}
  \caption{Evaluation of global acyclic constraint term $h(\theta^{(1)})$ and local causal graph}
  \label{tabel2}
\end{table}

\subsection{Results on a real dataset}
We evaluate FedCausal on a protein signaling network based on expression levels of proteins and phospholipid given by \cite{sachs2005causal}, which is generally accepted by the biology community. There are a total of $n=7466$ samples, $d=11$ variables, and $20$ edges in the ground truth DAG of this dataset (Sachs). We randomly select $n=7460$ samples and evenly distribute them to 10 clients to mimic scattered data. The results are shown in Table \ref{table3}.

\begin{table}[htbp]
\centering
\begin{tabular}{c| l| r r r r}
\hline
                    &           & TPR$\uparrow$ & FDR$\downarrow$   & SHD$\downarrow$   & NNZ\\
\hline
\multirow{2}{*}{CP} &PC         & \bf{50.00}    & 61.54             & 24                & 26 \\
                    &GES        & 40.00         & 78.95             & 31                & 38 \\
\hline
\multirow{6}{*}{DAG}&FedCausal  & \bf{45.00}    & \bf{52.63}        & \bf{15}           & 19 \\
                    &NO-ALL     & \bf{45.00}    & 57.14             & 18                & 21\\
                    &NO-w/oAcy  & 40.00         & 69.23             & 24                & 26 \\
                    &NO-Avg     & 40.00         & 70.37             & 25                & 27 \\
                    &FedDAG     & 40.00         & 71.43             & 26                & 28 \\
                    &NO-ADMM    & 20.00         & 80.00             & 29                & 20 \\
\hline
\end{tabular}
\caption{Results on the Sachs dataset. DAG and CP refer to algorithms that identify directed acyclic graph and complete partially directed acyclic graph, respectively.}
\label{table3}
\end{table}

FedCausal achieves the best and most reliable results. PC and GES, as centralized and typical causal discovery algorithms, both show good TPR but high SHD, because they can only identify CPDAGs. NO-ALL also uses all data to identify DAGs, so its results are better than most federated approaches. Federated methods NO-w/oAcy, NO-Avg, FedDAG and NO-ADMM are limited by decentralized data and are outperformed by centralized PC and NO-ALL. The performance of FedCausal is not only better than other federated methods, but even better than NO-ALL, which may be owed to the effective interaction between the clients and the server powered by our adaptive optimization.

\section{Conclusion}
This paper introduces a federated approach (FedCausal) to learn a unified global causal graph from decentralized non-iid data. FedCausal uses an explainable and adaptive optimization process to coordinate clients to optimize the local causal graphs based on clients' data and to learn the global graph with ensured acyclicity. Our analysis shows that the optimization objective of FedCausal under statistically homogeneous data is consistent with that of causal discovery algorithms for centralized data, and FedCausal can flexibly learn DAGs from decentralized heterogeneous data. Experimental results confirm its effectiveness, generality and reliability on iid, and non-iid data.

%%%%%%%%%%%%%%%%%%%%%%%%%%%%%%%%%%%%%%%%%%%%%%%%%%%%%%%%%%%%

\appendix
\setcounter{table}{0}
\setcounter{figure}{0}
\renewcommand{\thefigure}{S\arabic{figure}}
\renewcommand{\thetable}{S\arabic{table}}

\newpage

\twocolumn[
\begin{@twocolumnfalse}
	\section*{\centering{Federated Causality Learning with Explainable Adaptive Optimization \\
 Supplementary File \\[25pt]}}
\end{@twocolumnfalse}
]

\section{Experimental setups}

For all the federated baselines in the experiment, we used the default hyperparameters provided in the original papers or public codes. FedDAG needs to setup $\rho_{init}$ and $\beta$ to accommodate different numbers of nodes. $\rho_{init}$ is the initial penalty term coefficients in augmented lagrange method and $\beta$ is the update step size of $\rho_{init}$. The settings of $\rho_{init}$ and $\beta$ on 10, 20, 40 nodes are shown in Table \ref{tableS1}:

\begin{table}[htbp]
\centering
\vskip 0.15in
\begin{tabular}{c|r r r r}
\hline
                & 10 nodes          & 20 nodes          & 40 node   \\
\hline
$\rho_{init}$   & $6\times10^{-3}$  & $1\times10^{-5}$  & $1\times10^{-11}$\\
$\beta$         & 10                & 20                & 120              \\
\hline
\end{tabular}
\caption{$\rho_{init}$ and $\beta$ of FedDAG on simulated data}
\label{tableS1}
\end{table}
For all the NOTEARS based baselines in the experiment, we used the default hyperparameters of NOTEARS for linear data and the default htperparameters of NOTEARS-MLP for non-parametric data. For our FedCausal, we list all hyperparameters in Table \ref{tableS2}:

\begin{table}[htbp]
\scriptsize
\centering
\begin{tabular}{l|r r r r r r}
\hline
Parameters  & $\alpha_{init}$   & $\rho_{init}$ & $h_{tol}$         & $\rho_{max}$      & $\gamma$  & $\beta$\\
\hline
Values      & 0                 & 1             & $1\times10^{-11}$ & $1\times10^{16}$  & 0.25      & 10\\
\hline
\end{tabular}
\caption{Parameters of FedCausal}
\label{tableS2}
\end{table}

FedCausal used the same hyperparameters for all number of nodes and data models in the experiment. FedCausal updates $\alpha$ by the formula $\alpha=\alpha+\rho h(\theta^{(1)})$. To make the update step of $\alpha$ large enough, we compute $h(\theta^{(1)})$ and update $\alpha$ using the parameter $\theta^{(1)}$ obtained from the weighted average of local models.

\section{Efficiency study of FedCausal and baselines}

\begin{figure}[ht]
\centering
\subfigbottomskip=2pt
\includegraphics[scale=0.25]{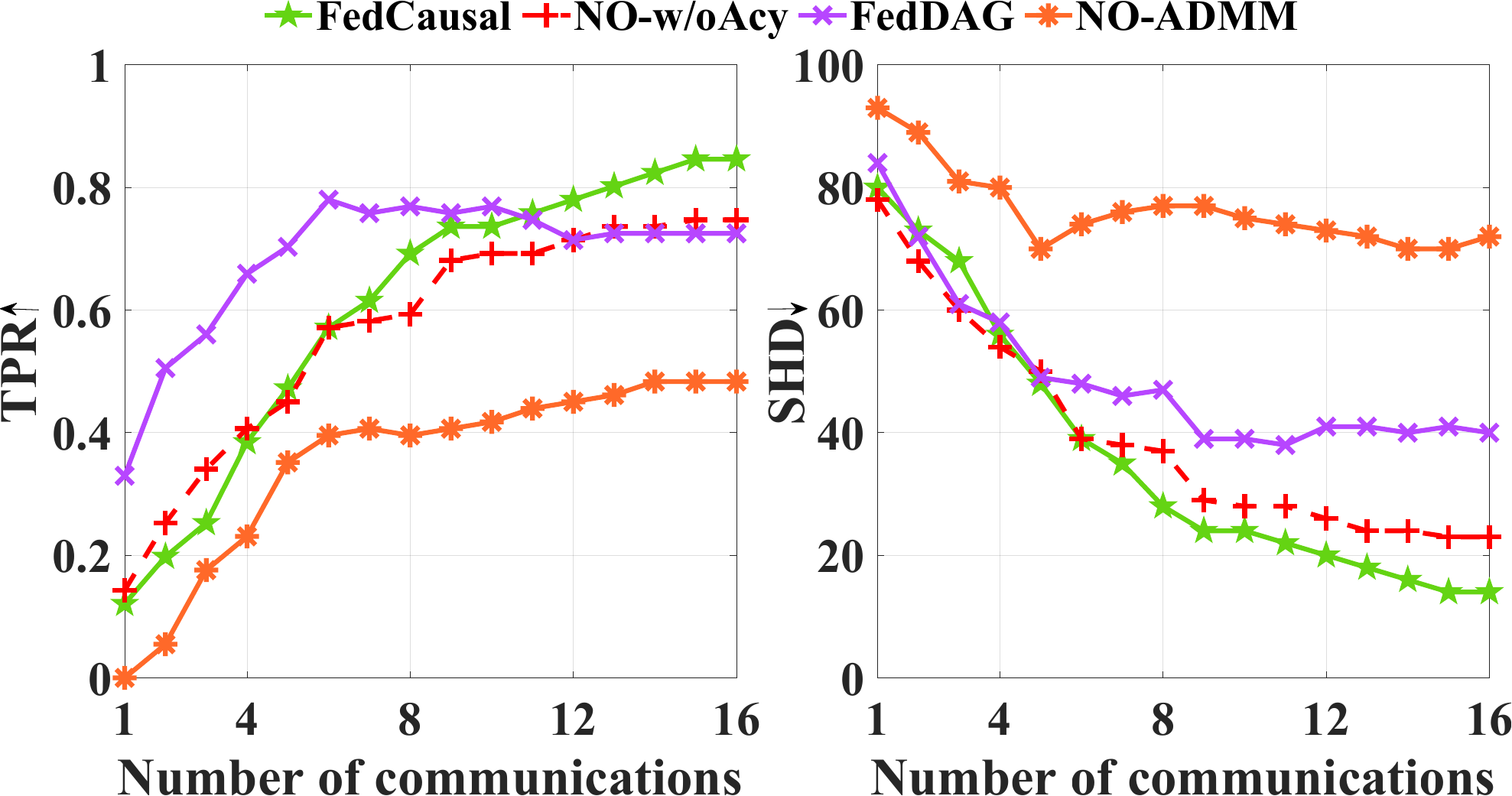} 
\caption{The performance of global graph vs. the number of communications}
\label{figS1}
\end{figure}

In this experiment, we set 10 clients, 200 samples for each client and graphs with 40 variables. We fixed the number of communications to 16 (FedCausal and baselines meet the stop condition after 16 communications) and record the metrics of the global graphs obtained by FedCausal and baselines after each communication on non-iid data. Figure \ref{figS1} shows the results of FedCausal and baselines.

FedCausal achieves the superior final results to other federated baselines. NO-ADMM's communication efficiency and found causal graphs are the worst, because it does not constrain the acyclicity and sparsity on the clients, resulting in the overfitting local models and failing to learn an accurate causal graph on the server. FedDAG has a rapidly improving TPR in the first few communications, but its performance decreases in the later communications. We believe that the mask mechanism adopted by FedDAG limits the update of the model and is harmful to the performance improvement of FedDAG, so FedDAG finally loses to FedCausal and NO-w/oAcy. Since the global acyclic constraint tends to make the global graph more sparse, FedCausal gets a slightly lower TPR with fewer edges than NO-w/oAcy at the beginning. However, the adaptive acyclic optimization enables FedCausal exchange more correct and effective causal information between the server and clients, so the TPR and SHD of FedCausal improve faster than those of NO-w/oAcy with the number of communications. After 16 communications, FedCausal outputs a more accurate global causal graph than the ones given by other baselines.

\section{Study of different local sample sizes}
To further study the robustness of FedCausal to different local sample sizes, we set 10 clients, graphs with 40 variables and 2000 samples. We divided all 2000 samples into 40 packages, each containing 50 samples. We then randomly distribute these 40 packages to these 10 clients (each client holds at least one package) to simulate different local sample sizes. We compare FedCausal with other baselines on non-iid data and record the results in Table \ref{tableS3}. The results of FedCausal with respect to the same number of client samples (200) are also reported for reference.

\begin{table}[htbp]
\centering
  \begin{tabular}{l| r| r| r}
    \hline
    40 nodes    &TPR$\uparrow$      &FDR$\downarrow$    &SHD$\downarrow$    \\
    \hline
    FedCausal(Even)   &\bf{84.9$\pm$4.7}  &\bf{0.9$\pm$1.2}   &\bf{13.3$\pm$4.8}\\
    \hline
    FedCausal   &\bf{86.0$\pm$3.8}  &\bf{2.5$\pm$2.8}   &\bf{13.6$\pm$4.6}\\
    NO-w/oAcy   &78.8$\pm$12.5      &\bf{3.9$\pm$3.2}   &20.7$\pm$11.2\\
    FedDAG      &72.5$\pm$7.7       &23.8$\pm$5.1       &36.2$\pm$6.6\\
    NO-ALL      &\bf{85.7$\pm$7.4}  &84.2$\pm$1.3       &389.9$\pm$70.3\\
    NO-Avg      &\bf{96.3$\pm$4.5}  &53.7$\pm$11.2      &107.1$\pm$45.9\\
    \hline
  \end{tabular}
  \caption{Results on clients with different sample sizes.}
  \label{tableS3}
\end{table}

We observe that FedCausal achieves the best results with different local sample sizes, it gets higher TPR than most of the baselines and the lowest FDR and SHD. NO-ALL and NO-Avg also have high TPR, but their results are not as reliable as FedCausal because their FDR and SHD are too poor.

\section{Study of the number of clients}
\begin{figure}[ht]
\centering
\subfigbottomskip=2pt
\includegraphics[scale=0.16]{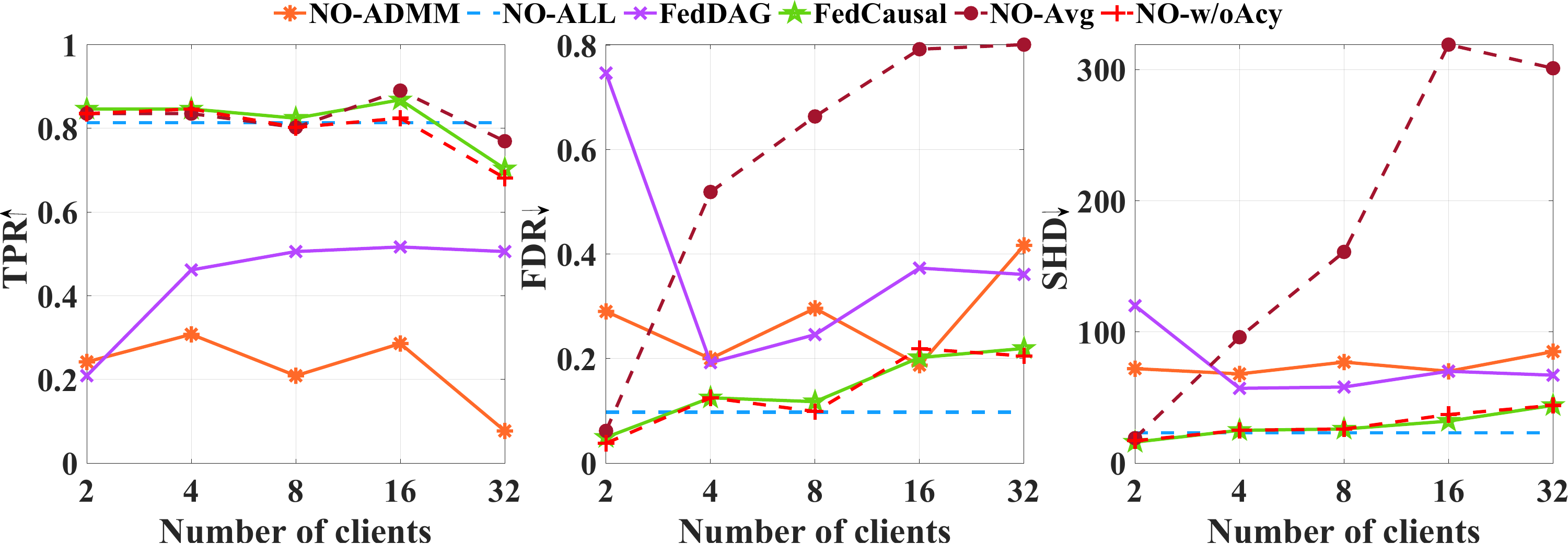} 
\caption{The global results and the number of communication}
\label{figS2}
\end{figure}

We also conduct experiments to investigate the impact of the number of clients. To reduce mixed factors, the experiments are performed on iid data with 40 variables and 2048 samplesm, where each client holds $\{1024,512,256,128,64\}$ samples for $\{2,4,8,16,32\}$ clients, respectively. Figure \ref{figS2} shows the results of FedCausal and other baselines.

NO-ALL learns causal graphs using all samples, so its results are not impacted by the number of clients and can be took as a benchmark. The results of FedCausal and NO-w/oAcy are very close to NO-ALL and are relatively stable under different numbers of clients. Their performance slightly degrades with 32 clients, this is because the local sample size is too small to effectively learn the local causal graph. NO-Avg achieves similar results to NO-ALL with 2 clients, but its FDR and SHD increase dramatically as the number of clients increases. This is because NO-Avg independently learns causal graphs from each client and does not aggregate local models, and with the increase of clients, the limited local sample size is too insufficient to learn accurate causal graphs. FedDAG and NO-ADMM lose to FedCausal on all metrics. It should be noted that FedDAG is very poor on 2 clients. This is because FedDAG aggregates binary causal graph (0-1 matrix), and it is difficult to aggregate valid results when there are only two local graphs.

%%%%%%%%%%%%%%%%%%%%%%%%%%%%%%%%%%%%%%%%%%%%%%%%%%%%%%%%%%%%


\begin{thebibliography}{31}
\providecommand{\natexlab}[1]{#1}

\bibitem[{Bernstein et~al.(2020)Bernstein, Saeed, Squires, and Uhler}]{bernstein2020ordering}
Bernstein, D.; Saeed, B.; Squires, C.; and Uhler, C. 2020.
\newblock Ordering-based causal structure learning in the presence of latent variables.
\newblock In \emph{AISTATS}, 4098--4108.

\bibitem[{B{\"u}hlmann, Peters, and Ernest(2014)}]{buhlmann2014cam}
B{\"u}hlmann, P.; Peters, J.; and Ernest, J. 2014.
\newblock CAM: Causal additive models, high-dimensional order search and penalized regression.
\newblock \emph{The Annals of Statistics}, 42(6): 2526--2556.

\bibitem[{Byrd et~al.(1995)Byrd, Lu, Nocedal, and Zhu}]{byrd1995limited}
Byrd, R.~H.; Lu, P.; Nocedal, J.; and Zhu, C. 1995.
\newblock A limited memory algorithm for bound constrained optimization.
\newblock \emph{SIAM Journal on Scientific Computing}, 16(5): 1190--1208.

\bibitem[{Chai et~al.(2020)Chai, Wang, Chen, and Yang}]{chai2020secure}
Chai, D.; Wang, L.; Chen, K.; and Yang, Q. 2020.
\newblock Secure federated matrix factorization.
\newblock \emph{IEEE Intelligent Systems}, 36(5): 11--20.

\bibitem[{Chickering(2002)}]{chickering2002optimal}
Chickering, D.~M. 2002.
\newblock Optimal structure identification with greedy search.
\newblock \emph{JMLR}, 3(11): 507--554.

\bibitem[{Gao et~al.(2023)Gao, Chen, Shen, Liu, Gong, and Bondell}]{gao2023feddag}
Gao, E.; Chen, J.; Shen, L.; Liu, T.; Gong, M.; and Bondell, H. 2023.
\newblock Fed{DAG}: Federated {DAG} Structure Learning.
\newblock \emph{TMLR}.

\bibitem[{Gou, Jun, and Zhao(2007)}]{gou2007learning}
Gou, K.~X.; Jun, G.~X.; and Zhao, Z. 2007.
\newblock Learning Bayesian network structure from distributed homogeneous data.
\newblock In \emph{In ACIS International Conference on Software Engineering, Artificial Intelligence, Networking, and Parallel/Distributed Computing}, volume~3, 250--254. IEEE.

\bibitem[{He et~al.(2021)He, Cui, Shen, Xu, Liu, and Jiang}]{he2021daring}
He, Y.; Cui, P.; Shen, Z.; Xu, R.; Liu, F.; and Jiang, Y. 2021.
\newblock Daring: Differentiable causal discovery with residual independence.
\newblock In \emph{KDD}, 596--605.

\bibitem[{Koller and Friedman(2009)}]{koller2009probabilistic}
Koller, D.; and Friedman, N. 2009.
\newblock \emph{Probabilistic graphical models: principles and techniques}.
\newblock MIT press.

\bibitem[{Lachapelle et~al.(2019)Lachapelle, Brouillard, Deleu, and Lacoste-Julien}]{lachapelle2019gradient}
Lachapelle, S.; Brouillard, P.; Deleu, T.; and Lacoste-Julien, S. 2019.
\newblock Gradient-Based Neural DAG Learning.
\newblock In \emph{ICLR}.

\bibitem[{Liang et~al.(2023)Liang, Wang, Yu, Guo, Domeniconi, and Guo}]{liang2023directed}
Liang, J.; Wang, J.; Yu, G.; Guo, W.; Domeniconi, C.; and Guo, M. 2023.
\newblock Directed Acyclic Graph Learning on Attributed Heterogeneous Network.
\newblock \emph{IEEE Transactions on Knowledge and Data Engineering}.

\bibitem[{McMahan et~al.(2017)McMahan, Moore, Ramage, Hampson, and y~Arcas}]{mcmahan2017communication}
McMahan, B.; Moore, E.; Ramage, D.; Hampson, S.; and y~Arcas, B.~A. 2017.
\newblock Communication-efficient learning of deep networks from decentralized data.
\newblock In \emph{AISTATS}, 1273--1282.

\bibitem[{Na and Yang(2010)}]{na2010distributed}
Na, Y.; and Yang, J. 2010.
\newblock Distributed Bayesian network structure learning.
\newblock In \emph{ISIE}, 1607--1611.

\bibitem[{Nemirovsky(1999)}]{nemirovsky1999optimization}
Nemirovsky, A. 1999.
\newblock Optimization II. Numerical methods for nonlinear continuous optimization.

\bibitem[{Ng and Zhang(2022)}]{ng2022towards}
Ng, I.; and Zhang, K. 2022.
\newblock Towards federated bayesian network structure learning with continuous optimization.
\newblock In \emph{AISTATS}, 8095--8111.

\bibitem[{Ng et~al.(2022)Ng, Zhu, Fang, Li, Chen, and Wang}]{ng2022masked}
Ng, I.; Zhu, S.; Fang, Z.; Li, H.; Chen, Z.; and Wang, J. 2022.
\newblock Masked gradient-based causal structure learning.
\newblock In \emph{SDM}, 424--432.

\bibitem[{Pearl(2009)}]{pearl2009causality}
Pearl, J. 2009.
\newblock \emph{Causality}.
\newblock Cambridge university press.

\bibitem[{Peters, Janzing, and Sch{\"o}lkopf(2017)}]{peters2017elements}
Peters, J.; Janzing, D.; and Sch{\"o}lkopf, B. 2017.
\newblock \emph{Elements of causal inference: foundations and learning algorithms}.
\newblock MIT Press.

\bibitem[{Sachs et~al.(2005)Sachs, Perez, Pe'er, Lauffenburger, and Nolan}]{sachs2005causal}
Sachs, K.; Perez, O.; Pe'er, D.; Lauffenburger, D.~A.; and Nolan, G.~P. 2005.
\newblock Causal protein-signaling networks derived from multiparameter single-cell data.
\newblock \emph{Science}, 308(5721): 523--529.

\bibitem[{Shimizu(2012)}]{shimizu2012joint}
Shimizu, S. 2012.
\newblock Joint estimation of linear non-Gaussian acyclic models.
\newblock \emph{Neurocomputing}, 81: 104--107.

\bibitem[{Spirtes et~al.(2000)Spirtes, Glymour, Scheines, and Heckerman}]{spirtes2000causation}
Spirtes, P.; Glymour, C.~N.; Scheines, R.; and Heckerman, D. 2000.
\newblock \emph{Causation, prediction, and search}.
\newblock MIT Press.

\bibitem[{Spirtes, Meek, and Richardson(2013)}]{spirtes2013causal}
Spirtes, P.~L.; Meek, C.; and Richardson, T.~S. 2013.
\newblock Causal inference in the presence of latent variables and selection bias.
\newblock \emph{arXiv preprint arXiv:1302.4983}.

\bibitem[{Tillman, Danks, and Glymour(2008)}]{danks2008integrating}
Tillman, R.~E.; Danks, D.; and Glymour, C. 2008.
\newblock Integrating locally learned causal structures with overlapping variables.
\newblock In \emph{NeurIPS}, 1665--1672.

\bibitem[{Triantafillou and Tsamardinos(2015)}]{triantafillou2015constraint}
Triantafillou, S.; and Tsamardinos, I. 2015.
\newblock Constraint-based causal discovery from multiple interventions over overlapping variable sets.
\newblock \emph{JMLR}, 16(1): 2147--2205.

\bibitem[{Tsamardinos, Brown, and Aliferis(2006)}]{tsamardinos2006max}
Tsamardinos, I.; Brown, L.~E.; and Aliferis, C.~F. 2006.
\newblock The max-min hill-climbing Bayesian network structure learning algorithm.
\newblock \emph{Mach. Learn.}, 65(1): 31--78.

\bibitem[{Xiong et~al.(2021)Xiong, Koenecke, Powell, Shen, Vogelstein, and Athey}]{xiong2021federated}
Xiong, R.; Koenecke, A.; Powell, M.; Shen, Z.; Vogelstein, J.~T.; and Athey, S. 2021.
\newblock Federated causal inference in heterogeneous observational data.
\newblock \emph{arXiv preprint arXiv:2107.11732}.

\bibitem[{Ye, Amini, and Zhou(2022)}]{ye2022distributed}
Ye, Q.; Amini, A.~A.; and Zhou, Q. 2022.
\newblock Distributed Learning of Generalized Linear Causal Networks.
\newblock \emph{arXiv preprint arXiv:2201.09194}.

\bibitem[{Yu et~al.(2019)Yu, Chen, Gao, and Yu}]{yu2019dag}
Yu, Y.; Chen, J.; Gao, T.; and Yu, M. 2019.
\newblock DAG-GNN: DAG structure learning with graph neural networks.
\newblock In \emph{ICML}, 7154--7163.

\bibitem[{Yuan(2011)}]{yuan2011identifiability}
Yuan, M. 2011.
\newblock On the identifiability of additive index models.
\newblock \emph{Statistica Sinica}, 1901--1911.

\bibitem[{Zheng et~al.(2018)Zheng, Aragam, Ravikumar, and Xing}]{zheng2018dags}
Zheng, X.; Aragam, B.; Ravikumar, P.; and Xing, E.~P. 2018.
\newblock DAGs with NO TEARS: continuous optimization for structure learning.
\newblock In \emph{NeurIPS}, 9492--9503.

\bibitem[{Zheng et~al.(2020)Zheng, Dan, Aragam, Ravikumar, and Xing}]{zheng2020learning}
Zheng, X.; Dan, C.; Aragam, B.; Ravikumar, P.; and Xing, E. 2020.
\newblock Learning sparse nonparametric dags.
\newblock In \emph{AISTATS}, 3414--3425.

\end{thebibliography}
\end{document}